\documentclass[11pt, a4paper, onecolumn, copyright, goog]{google}

\usepackage[authoryear, sort&compress, round]{natbib}
\usepackage{xspace}
\usepackage{amsmath}
\usepackage{caption}
\usepackage{algorithm}
\usepackage{algorithmic}
\usepackage{multirow}
\usepackage{multicol}
\usepackage{graphicx}
\usepackage{wrapfig}
\usepackage{float}
\usepackage{listings}
\usepackage{xcolor}
\usepackage{enumitem}

\newcommand{\ourmethod}{MILO\xspace}
\newcommand{\minisection}[1]{\vspace{0.05in}\noindent {\bf #1}}

\keywords{In-Context Learning, Large Language Model, KV Cache Compression}

\uselogo{} 

\title{\ourmethod: Efficient Many-shot In-Context Learning with Block-wise Low-rank Compression}

\correspondingauthor{ypzhao96@gmail.com, tiantan@google.com}

\reportnumber{} 

\author[1,2]{Youpeng Zhao}
\author[1]{Tian Tan}
\author[1]{Liqian Peng}
\author[2]{Jun Wang}
\author[1]{Alec Go}

\affil[1]{\thepa{}{}}
\affil[2]{University of Central Florida}

\begin{abstract}
Many-shot in-context learning (ICL) enables large language models (LLMs) to adapt to complex tasks by conditioning on thousands of demonstration examples, but this paradigm shifts the inference efficiency bottleneck to the key-value (KV) cache memory. 
Due to the linear scaling behavior of the KV cache, storing these intermediate tensors has become a paramount challenge for both online serving and on-device deployment.
To address this issue, we propose a novel compression framework, termed \textbf{\ourmethod}, that exploits the low-rank redundancy inherent in many-shot contexts. 
Specifically, \ourmethod features a block-wise low-rank compression strategy that compresses the KV cache at the block granularity, where each block contains multiple many-shot examples.
Furthermore, to handle the heterogeneous context density across different blocks, \ourmethod dynamically allocates rank budgets based on the information entropy, preserving the fidelity of critical blocks while aggressively compressing redundant ones. 
Experimental results on Qwen2.5 models demonstrate that our method achieves up to 50\% reduction in KV cache memory and 1.8$\times$ throughput improvement, with negligible performance degradation on classification and reasoning benchmarks, significantly outperforming prior baselines.
\end{abstract}

\begin{document}

\maketitle

\section{Introduction}

In-context learning (ICL)~\citep{gpt3,manyshoticl1,manyshoticl2,manyshoticl3} is a powerful capability of large language models (LLMs)~\cite{llama,gpt3,qwen2} that enables the model to learn a new task at inference time by conditioning on a few demonstration examples or `shots' provided directly in the prompt, without any parameter updates. 
Traditionally, it has been limited to "few-shot" learning due to the small context windows of prior models~\cite{gpt3}. 
However, with the recent expansion of model context windows to a million tokens or more, a new paradigm of `many-shot ICL' has emerged~\cite{manyshoticl1,manyshoticl2,manyshoticl3,dbsa}. 
Through augmenting prompts with hundreds or even thousands of examples, many-shot ICL has been shown to significantly boost performance on complex downstream tasks, often approaching the accuracy of fully fine-tuned models~\cite{manyshoticl1,manyshoticl2}. 

\noindent Despite its effectiveness across a wide range of tasks, many-shot ICL introduces significant efficiency challenges for practical deployment~\cite{iclsurvey,dbsa,tcm}.
Long-context inputs place significantly higher memory and compute requirements for inference due to the quadratic complexity of the attention operation and the existence of key-value (KV) cache~\cite{fairseq}.
To maintain high performance and low latency, many-shot ICL often requires careful and query-dependent example selection to construct proper inputs~\cite{reticl,tcm}.
Recent works have attempted to minimize the inference costs by pre-encoding the KV cache of many-shot examples~\cite{dbsa,adapshot}.
Notably, DBSA applies dynamic sparse attention to alleviate the compute burden, and Adapshot further introduces semantic-aware KV cache reuse~\cite{adapshot}.
However, none of the prior works discuss the storage and memory efficiency aspects of the KV cache.
For instance, storing the KV cache for 90k examples necessitates an additional 11.1 GB for 8B dense models~\cite{dbsa}.
As the long-context capabilities are continuously expanded, with models like Gemini now supporting 10M tokens, the growing memory footprint for encoding prior examples makes many-shot ICL increasingly memory-bound and impractical for serving with stringent service-level objectives (SLOs).

\noindent In this work, we present an efficient framework, termed \textbf{\ourmethod}, to address the above-mentioned KV cache challenges in many-shot ICL.
\textbf{Our key idea is to exploit the observed low-rank nature of the KV cache of pre-encoded many-shot examples.
This low-rank redundancy arises because many examples share similar semantic features, leading to highly correlated representations within the KV cache in latent space.}
By storing the KV cache in low-dimensional space, \ourmethod can achieve up to 50\% storage cost reduction and 1.8$\times$ speedup against prior works with minimal accuracy degradation.
However, achieving this performance requires solving two unique challenges.

\noindent First, it is essential to choose the appropriate granularity for low-rank compression to achieve a balance between memory efficiency and model performance. 
Compression itself is a lossy process, where a naive, global approach risks losing fine-grained information across different examples and, most importantly, induces redundant computation overhead during decompression~\cite{palu}.
Conversely, attempting to compress on an example-by-example basis fails to capture the global semantics and results in a significant performance drop due to accumulated information loss from each example~\cite{ShadowKV}.
To resolve this trade-off, \ourmethod employs a block-wise low-rank compression (BLC) method.
Specifically, the KV caches are compressed at the granularity of a block, where each block contains multiple many-shot examples.
Such a granularity can strike the best trade-offs between model accuracy and inference efficiency, thereby improving the end-to-end quality of service (QoS).

\noindent A second challenge lies in the fact that the information density within the KV cache is highly heterogeneous across the blocks. 
While some blocks focus on local semantic patterns that are highly redundant, others often retain critical information required for tasks like complex chain-of-thought (CoT) reasoning~\cite{cot1,cot2} and test-time scaling~\cite{s1,tts1,tts2,tts3}. 
Applying a uniform compression rank across the entire cache is therefore often suboptimal. 
To address this issue, we propose a dynamic rank adaptation mechanism that allocates higher rank budgets based on the entropy of each block and aggressively compresses redundant ones.
This ensures that the compression algorithm can preserve the fidelity of retrieval capabilities while minimizing the aggregate KV cache memory footprint.
Furthermore, we design and implement a suite of system optimizations, e.g., customized kernels and CUDA graphs and streams, tailored to \ourmethod, to achieve real-world end-to-end benefits.

\noindent In summary, our contributions are as follows:
\begin{itemize}
    \item We identify the KV cache memory challenges in many-shot ICL and propose a new framework, termed \ourmethod, by exploiting the low-rank redundancy inherent in the KV cache during many-shot ICL.
    \item \ourmethod features block-wise low-rank compression (BLC) that balances reconstruction efficiency with model representation quality, and a dynamic rank adaptation mechanism to allocate ranks based on the information entropy.
    \item Evaluations on Qwen2.5 models demonstrate that our framework significantly reduces KV cache storage and improves inference latency with negligible accuracy degradation across a suite of standard ICL benchmarks.
\end{itemize}

\section{Background}
\label{sec:bg}
\minisection{Many-Shot ICL.} 
The paradigm of in-context learning (ICL) was first popularized by GPT-3~\cite{gpt3}, demonstrating that LLMs can adapt to new tasks by conditioning on a few demonstration examples without parameter updates. 
While early work focused on standard "few-shot" settings (typically 1$\sim$8 examples) due to limited context windows (e.g., 2k or 4k tokens), the recent advent of long-context models—capable of processing up to 10M tokens—has unlocked the potential for "many-shot" ICL.
Recent studies indicate that scaling the number of demonstrations from dozens to thousands yields monotonic performance improvements, rivaling the performance of fine-tuned models on complex reasoning and extraction tasks~\cite{manyshoticl1, manyshoticl2}. 

\minisection{Inference Optimization for Many-Shot ICL.}
Despite the promising results of many-shot ICL, this approach shifts the bottleneck from offline training to online inference, where the prompt length scales linearly with the number of shots.
The quadratic complexity of self-attention and, more critically, the linear growth of the key-value (KV) cache imposes severe memory and latency constraints~\cite{fairseq}. 
Several recent works have attempted to resolve such inference challenges.
Notably, DBSA explores pre-encoding mechanisms with dynamic sparse attention to reduce compute latency~\cite{dbsa}.
Adapshot proposes dynamic context budget allocation and KV cache reuse to further improve both accuracy performance and system throughput~\cite{adapshot}.
However, the challenge of managing the memory footprint for the cached intermediate states of thousands of examples remains an open problem that our work aims to address.

\minisection{Low-rank Compression.}
Several works have explored compressing LLMs in low-rank space. 
Prior efforts have generally focused on weight and activation compression using Fisher information~\cite{asvd,mc}.
Recently, as context length continues to scale up, several works aim to further compress KV cache for more efficient memory management~\cite{palu,EffectivelyCK,MatryoshkaKV,Eigen,ShadowKV,ck,lorc}.
Notably, Palu applies a low-rank projection matrix in zero-shot settings~\cite{palu}, and LoRC proposes inter-layer progressive compression~\cite{lorc}
~\cite{ck} explores compressing KV cache at the head level by converting multi-head attention (MHA) to group-query attention (GQA) with LoRA fine-tuning~\cite{lora}.
\textit{However, existing works focus on the compression within a limited context window and do not consider the low-rank nature across examples.
Our work aims to extend low-rank compression to many-shot settings and fill in the gap.}

\minisection{Sparse Attention and Quantization.} 
Another line of inference-time approaches is based on the observation that not all tokens are created equal, thus creating the opportunity of sparse attention. 
$H_2O$~\cite{h2o}, StreamingLLM~\cite{streamingllm}, and Scissorhands~\cite{Scissorhands}, aim to reduce memory by evicting tokens deemed less important, typically retaining the most important ones for token generation. 
On the system level, ALISA~\cite{alisa} employs a co-design approach to maximize throughput performance on resource-constrained platforms. 
DBSA applies a similar block sparse attention method for retrieval-based many-shot in-context learning, achieving competitive performance against fine-tuning baselines~\cite{dbsa}.
Quantization is also a standard method for reducing KV cache memory footprint.
Techniques like KIVI~\cite{kivi} and KVQuant~\cite{KVQuant} compress the KV cache by reducing numerical precision, storing entries in 4-bit or 2-bit formats. 
\textit{Orthogonal to these works, our method aims to compress the representation of the KV cache from the perspective of low-rank approximation, which can be combined with quantization and sparse attention for further gains.
}

\newpage 

\section{Observation}
\label{sec:ovbservation}
\begin{wrapfigure}[15]{r}{0.48\textwidth}
    \centering
    \vspace{-17pt}
    \includegraphics[width=0.48\textwidth]{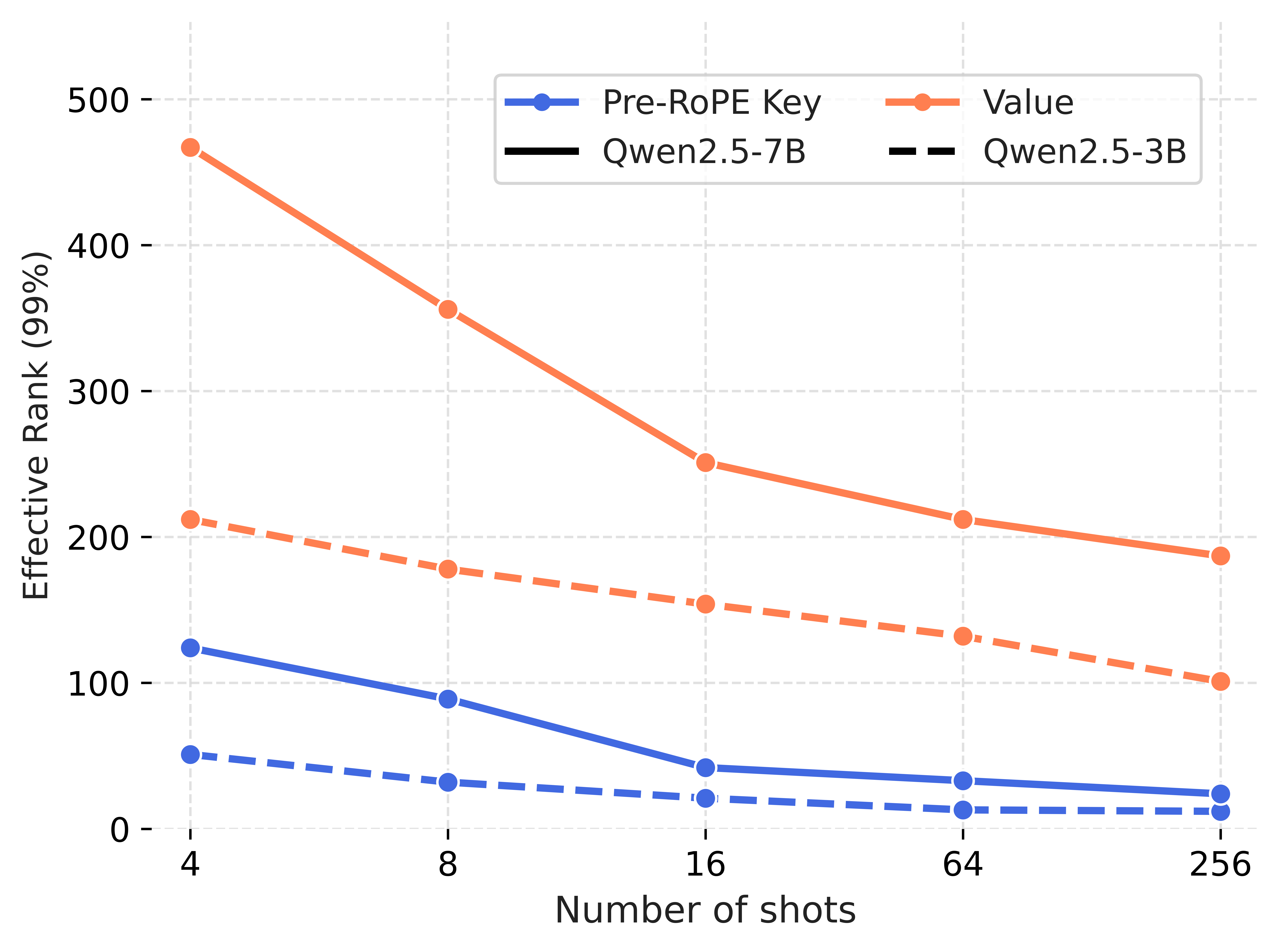}
    \vspace{-20pt}
    \caption{Top-1\% Effective rank of pre-RoPE key and value over different numbers of shots on the MathQA dataset.}
    \label{fig:kv_cache}
\end{wrapfigure}
\begin{figure*}[!t]
    \centering
    \includegraphics[width=\linewidth]{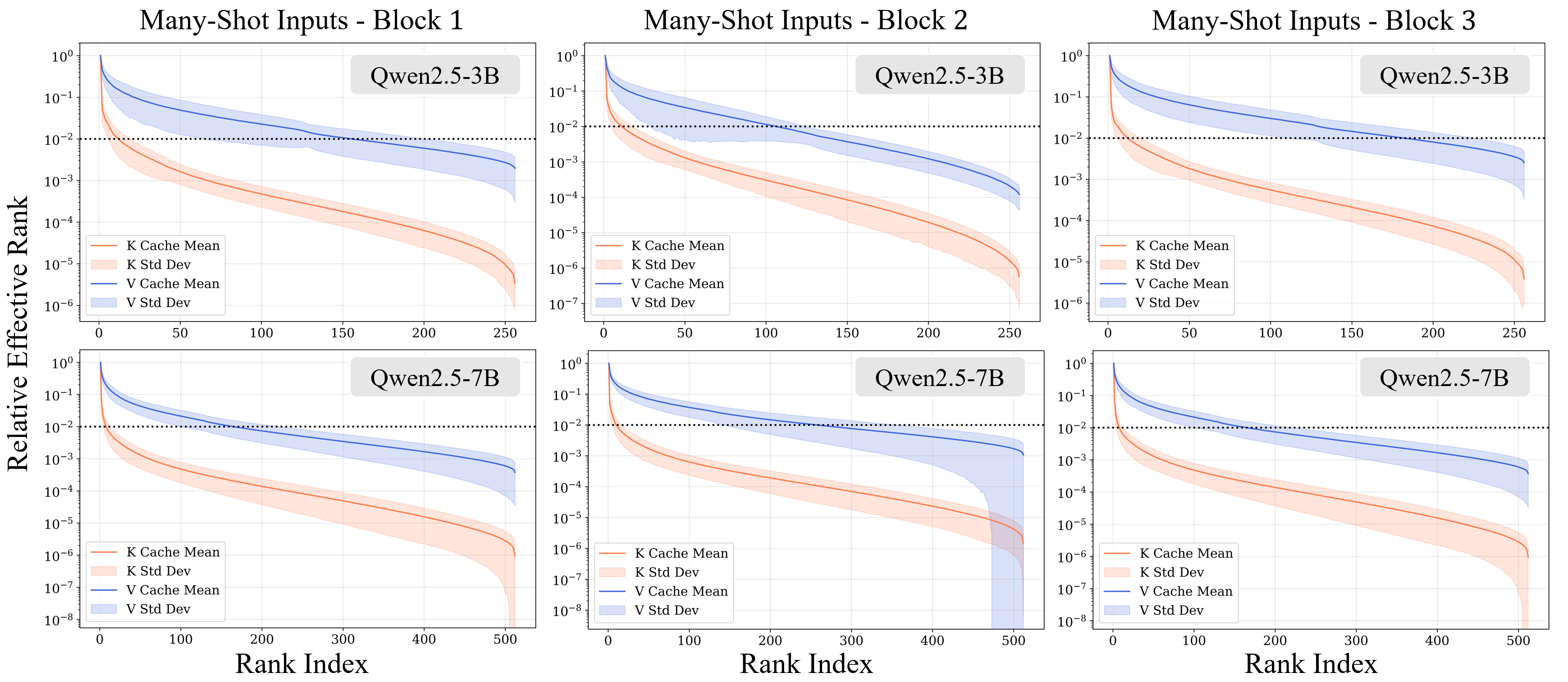}
    \vspace{-6mm}
    \caption{Observation of the low-rank characteristics of the key and value cache in Qwen2.5 models across different blocks of examples on MATHQA datasets.
    Here we sample three blocks, where each block contains 64 examples.
    Effective rank is defined as the square of singular values.
    We calculate the mean and std across all attention layers. 
    }
    \label{fig:insight}
\end{figure*}

To reduce memory footprint, recent works have explored the low-rank nature of the KV cache~\cite {palu,EffectivelyCK,MatryoshkaKV,Eigen,ShadowKV}.
However, these methods focus on data-dependent compression, which often demands additional training or finetuning to achieve competitive compression rates~\cite{palu,Eigen}.
Furthermore, these studies are often limited to zero- or few-shot learning settings with limited context lengths.
In our study, we aim to uncover the low-rank characteristics of the KV cache in many-shot ICL settings.
Figures~\ref {fig:kv_cache} and~\ref{fig:insight} visualize the singular value distributions of pre-RoPE key and value caches on Qwen2.5 models with many-shot inputs.

\noindent Here we have three key observations.
\underline{First}, the key cache exhibits a much stronger low-rank nature than the value cache in many-shot ICL settings, as shown in both Figure~\ref{fig:kv_cache} and ~\ref{fig:insight}.
This is consistent with prior findings that pre-RoPE keys are much better suited for low-rank compression~\cite{ShadowKV}.
\underline{Second}, counterintuitively, scaling the number of shots in ICL creates more KV storage, but simultaneously introduces more redundancy in low-rank space, as shown in Figure~\ref{fig:kv_cache}.
\underline{Third}, across different blocks and different layers, the low-rank behavior is heterogeneous, as indicated in Figure~\ref{fig:insight}.
These observations both motivate and validate our solution to exploit the low-rank nature of many-shot contexts.

\section{Methodology}
\label{sec:method}
In this section, we present our framework for efficient many-shot in-context learning. We first formulate the memory bottleneck problem in standard transformer attention. 
We then introduce block-wise low-rank compression, our approach for granular cache compression, and a dynamic rank allocation mechanism with efficient attention.
Next, we present some system optimization techniques for \ourmethod.

\subsection{Background}
Consider a transformer-based LLM with $L$ layers and $H$ attention heads per layer. 
In the many-shot setting, the model processes a sequence of inputs $X = \{x_1, \dots, x_N\}$, where $N$ represents a long context (e.g., $N > 10^5$). The standard self-attention mechanism for a specific head $h$ at layer $l$ computes the output $\mathbf{O}_h$ as:
\begin{equation}
    \mathbf{O}_{h} = \text{Attn}(\mathbf{Q_h}, \mathbf{K_h}, \mathbf{V_h}) =  \text{Softmax}\left(\frac{\mathbf{Q}_h \mathbf{K}_h^\top}{\sqrt{d}}\right) \mathbf{V}_h
\end{equation}
\noindent where $\mathbf{Q}_h, \mathbf{K}_h, \mathbf{V}_h \in \mathbb{R}^{N \times d}$ denote the Query, Key, and Value matrices, respectively, and $d$ is the head dimension.
The primary bottleneck in many-shot ICL is the storage of the intermediate key-value tensors, i.e., the KV cache, $(\mathbf{K}, \mathbf{V})$, which grows linearly with the sequence length $N$~\cite{fairseq}. 
The total memory footprint $\mathcal{M}$ required to store the KV cache in FP16 precision is calculated as:
\begin{equation}
    \mathcal{M} = 2 \cdot N \cdot L \cdot H \cdot d \cdot 2 
\end{equation}
As the number of shots continues to grow, $\mathcal{M}$ can often exceed the high-bandwidth memory (HBM) capacity of modern GPUs, forcing evictions or offloading that degrade performance. 
Our objective is to compress $(\mathbf{K}, \mathbf{V})$ into approximations $(\hat{\mathbf{K}},\hat{\mathbf{V}})$ such that the memory usage is minimized while preserving the model's accuracy performance.
\subsection{Block-wise Low-Rank Compression}
\label{subsec:BLC}
Global compression methods often treat the entire context matrix as a single entity, but this approach risks losing fine-grained details essential for many-shot ICL. 
It also induces redundant computation during decoding, as only a handful of many-shot examples are selected for the entire KV cache~\cite{dbsa}. 
On the other hand, compressing each example or token leads to improved decoding efficiency, but suffers significant performance degradation, as the information loss from each example accumulates~\cite{ShadowKV,asvd}.  
\begin{figure*}[!t]
    \centering
    \includegraphics[width=\linewidth]{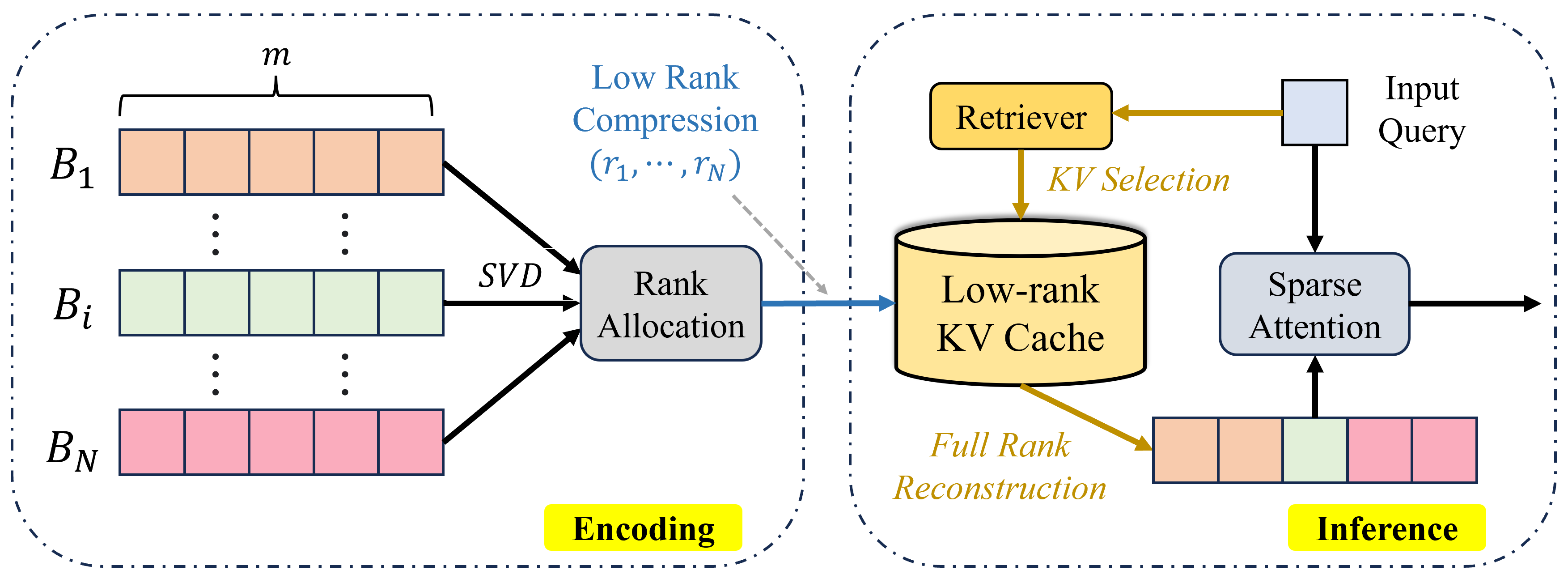}
    \vspace{-6mm}
    \caption{Overview of \ourmethod.
    During the encoding stage, the KV cache for many-shot contexts is split into $N$ blocks, with each block containing $m$ examples.
    \ourmethod first applies SVD to compute the appropriate rank for each block based on entropy analysis, then performs subsequent low-rank compression with rank budgets $(r_{1},...,r_{N})$.
    During inference, the input query goes through the retriever to conduct KV cache selection, where the selected compressed KV cache is reconstructed in parallel to perform subsequent sparse attention.
    }
    \label{fig:overview}
\end{figure*}
To address the above-mentioned limitations, we exploit the insights from Section~\ref{sec:ovbservation}, and propose a \textbf{block-wise low-rank compression}.
Specifically, we partition the KV cache into $N$ blocks, each containing multiple many-shot examples. 
Given a block of key-value pairs $\mathbf{K}_i, \mathbf{V}_i \in \mathbb{R}^{B \times d}$, where $B$ is the block size (total tokens across blocked examples) and $d$ is the head dimension, we apply low-rank approximation to each block independently. 
Given that key and value often exhibit different levels of low-rank, for each block $i$, we perform a decomposition for $\mathbf{K}_i$ and $\mathbf{V}_i$ separately:
\begin{equation}
    \mathbf{K}_i \approx \mathbf{A}_i^{\mathbf{k}} \mathbf{\Sigma}_i^{\mathbf{k}} ({\mathbf{B}_i^{\mathbf{k}}})^\top, \quad 
    \mathbf{V}_i \approx \mathbf{A}_i^{\mathbf{v}} \mathbf{\Sigma}_i^{\mathbf{v}} ({\mathbf{B}_i^{\mathbf{v}}})^\top
\end{equation}
where $\mathbf{A}_i \in \mathbb{R}^{B \times r_i}$, $\mathbf{\Sigma}_i \in \mathbb{R}^{r_i \times r_i}$ is a diagonal matrix of singular values, and $\mathbf{B}_i \in \mathbb{R}^{d \times r_i}$. 
$r_i$ denotes the rank allocated to block $i$, satisfying $r_i \ll \min(B, d)$.
Instead of storing $2Bd$ parameters per block, BLC stores the factorized matrices, reducing the storage cost to $(r_i^{\mathbf{k}}+r_i^{\mathbf{v}})(B + d)$. 
Assume a uniform block size $B$, the overall compression ratio $\rho$ is given by:
\begin{equation}
    \rho = \sum_{i=1}^{N}\frac{(r_i^{\mathbf{k}}+r_i^{\mathbf{v}})(B+d)}{2 \cdot N \cdot B \cdot d}
\end{equation}
This granularity allows us to capture local semantic dependencies while maintaining a high compression ratio.

\subsection{Dynamic Rank Allocation}
\label{sec:rank_adaptation}
As shown in Figure~\ref{fig:insight}, the information density in many-shot prompts is highly heterogeneous.
This echoes prior works that a uniform rank allocation for the KV cache at each layer is often suboptimal~\cite{palu,lorc}.
Here, we further demonstrate that this phenomenon also extends to many-shot ICL at the block level.
Specifically, in a typical many-shot setting, the context window is populated with numerous demonstration blocks, but different blocks exhibit substantially different spectral complexity and therefore require different representation capacities under compression.
Therefore, applying a uniform rank wastes representation capacity on highly redundant blocks while under-allocating rank to information-dense blocks. 

\noindent Here, we formulate rank selection as a global resource allocation problem under a fixed total rank budget and dynamically assign the rank $r_i$ for each block based on its corresponding entropy.
The notion of entropy is often used to measure such expressiveness and information density in neural networks~\cite{merino,mae,redunet,asvd}.
Specifically, for block $i$, we perform singular value decomposition (SVD) to obtain the singular values $\{\sigma_{1}, ..., \sigma_{d}\}$, where $\sigma_{1} \ge \sigma_{2} \ge ... \ge \sigma_{d}$.
Following prior works~\cite{merino,redunet}, we define the retained spectral information at rank $d$ as:
\begin{equation}
    \mathbf{H}_i(d) \triangleq  \sum_{j=1}^{d} \log (1+ \frac{\sigma_j^2}{\epsilon^2})
\end{equation}
where $\epsilon$ is a small constant for numerical stability. 
Given a total rank budget $R_{\mathrm{tot}}$, MILO allocates block-specific ranks $\{r_i\}_{i=1}^{N}$ by solving the following optimization problem:
\begin{equation}
\begin{aligned}
\max_{\{r_i\}}
\quad
\sum_{i=1}^{N} H_i(r_i), \quad
\text{s.t.}\quad
\sum_{i=1}^{N} r_i \le R_{\mathrm{tot}}, \quad 
r_{\min} \le r_i \le r_{\max}.
\end{aligned}
\label{eq:rank_budget}
\end{equation}

\noindent This formulation allocates more rank to blocks with slowly decaying
spectra while aggressively compressing redundant blocks.
Since $H_i(r)$ is additive over singular components, the allocation can
be efficiently computed using their marginal information gains. The gain
from increasing the rank of block $i$ from $r$ to $r+1$ is
\begin{equation}
\Delta H_{i,r} = H_i(r+1)-H_i(r) = \log \left(1+\frac{\sigma_{i,r+1}^{2}}{\epsilon^{2}}
\right).
\end{equation}

\noindent Starting from $r_i=r_{\min}$, we can greedily assign each remaining rank
unit to the block with the largest marginal gain until
$\sum_i r_i = R_{\mathrm{tot}}$. 
This produces a heterogeneous rank
allocation while strictly satisfying the global compression budget.

\subsection{\ourmethod Workflow}
\begin{minipage}[t]{0.45\textwidth}
Combining the above two proposed techniques, we present \ourmethod, a new framework for efficient many-shot ICL.
Figure~\ref{fig:overview} and Algorithm~\ref{alg:milo} further demonstrate the detailed workflow of our framework.
Here, for simplicity, we assume $\mathbf{K}$ and $\mathbf{V}$ are compressed to the same rank, while in practice they can be compressed to different ranks according to the entropy distribution.
It is divided into an encoding stage and an inference stage. 
During encoding, the KV cache for a many-shot context is partitioned into multiple blocks, each containing a specific number of examples. 
These blocks undergo SVD to determine an optimal rank allocation. 
By analyzing the entropy of each block, the system applies low-rank compression based on assigned rank budgets and stores the resulting data in a low-rank KV cache. 
In the subsequent inference stage, an input query is first evaluated by a retriever to perform KV selection from this compressed storage. 
The selected low-rank cache segments then undergo full-rank reconstruction in parallel. 
Finally, the reconstructed KV cache and the original input query are processed together through a sparse attention mechanism to complete the generation.
\end{minipage}%
\hfill 
\begin{minipage}[t]{0.52\textwidth}
\vspace{-18pt} 
\begin{algorithm}[H] 
   \caption{\ourmethod Workflow}
   \label{alg:milo}
\begin{algorithmic}
    \STATE {\bfseries Input:} Many-shot context $\mathcal{C}$ with $N$ blocks $\{B_1, B_2, \dots, B_N\}$, where each $B_i$ contains $m_i$ examples. 
    Low-rank Cache $LC$. 
    Input Query $\mathbf{Q}$.
    \STATE {\bfseries Output:} Attention outputs $\mathbf{O}$
    \STATE \textit{Stage 1: Encoding}
    \FOR{$i == 1$ to $N$}
        \STATE \emph{\# Perform SVD}
       \STATE $\mathbf{A}, \mathbf{\Sigma}, \mathbf{B}^\top = \text{SVD}((\mathbf{K}_i, \mathbf{V}_i))$
       \STATE \emph{// Get Optimal Rank $r_i$}
       \STATE $r_i = \text{argmin}_{r_{i}} \mathbf{H}(\cdot)$
       \STATE \emph{// Compress and Store}
       \STATE $LC.Store(\mathbf{A}[:r_i], \mathbf{\Sigma}[:r_i], \mathbf{B}[:r_i])$
    \ENDFOR
    \STATE \textit{Stage 2: Inference}
    \STATE \emph{\# Process Query to the Retriever}
    \STATE $(k_1, k_2, ..., k_N)=Retriver(\mathbf{Q})$
    \FOR{$j == k_1$ to $k_N$}
    \STATE \emph{// Retrieve Selected Block}
    \STATE $\mathbf{A}_j, \mathbf{\Sigma}_j, {\mathbf{B}_j}= LC.Retreive(j)$
    \STATE \emph{// Full-rank Reconstruction}
    \STATE $(\mathbf{K}_i, \mathbf{V}_i) \approx \mathbf{A}_i \mathbf{\Sigma}_i {\mathbf{B}_i}^\top$
    \ENDFOR
    \STATE $\mathcal{O} = \text{Attn}(\mathbf{Q}, [\mathbf{K}_{k_1},...\mathbf{K}_{k_N}], [\mathbf{V}_{k_1},...\mathbf{V}_{k_N}])$
    \STATE \textbf{return} $\mathcal{O}$
\end{algorithmic}
\end{algorithm}
\end{minipage}

\subsection{System Optimizations}
\label{sec:system_optimization}

To translate \ourmethod's memory savings into end-to-end speedup, we optimize the reconstruction path with fused Triton kernels, CUDA streams, and CUDA Graphs.

\minisection{Fused Low-Rank Reconstruction.}
A naive PyTorch-based implementation requires separate kernels for loading low-rank
factors, applying singular-value scaling, reconstructing KV tensors, and
writing them back to global memory. 
We instead implement a fused Triton
kernel that performs cache retrieval, scaling, and reconstruction in a
single pass. 
The kernel tiles the token and head dimensions, computes the reconstructed keys and values in registers and on-chip SRAM, and writes
them directly to the KV buffer consumed by attention. 
This reduces both intermediate HBM traffic and intermediate kernel-launch overhead.

\minisection{Parallel Reconstruction with CUDA Streams.}
Rather than reconstructing them sequentially on the default stream, \ourmethod dispatches independent blocks to multiple CUDA streams to enable block-level parallelism and synchronizes only before sparse attention. 
This improves GPU utilization for small reconstruction kernels and reduces reconstruction latency on the critical path.

\minisection{CUDA Graph Execution.}
We further use CUDA Graphs to reduce CPU launch overhead for repeated
inference. Frequently occurring reconstruction and attention execution
paths are captured and replayed for compatible request configurations,
avoiding repeated kernel-launch overhead for short GPU operations.

\section{Evaluation}
\label{sec:eval}
\subsection{Experimental Setup}
\minisection{Models and Datasets.} 
We conduct experiments on Qwen family models, which are widely used for long-context understanding and reasoning tasks in real-world deployment.
Specifically, we choose Qwen2.5-3B and Qwen2.5-7B~\cite{qwen25}. 
In terms of datasets, we focus on natural language understanding and mathematical reasoning tasks, including Banking77~\cite{banking77}, Clinic150~\cite{clinic150}, TREC~\cite{trec},  NLU~\cite{nlu}, MathQA~\cite{mathqa}, and SVAMP~\cite{svamp}.
We report model performance as the average accuracy scores of the above tasks.
For system performance, we report both the memory and throughput results in an end-to-end fashion.
All models are running in half-precision (FP16).

\minisection{Baselines.} 
We compare our method against standard full caching and state-of-the-art low-rank methods:
\begin{itemize}[nosep]
    \item \textbf{Full Cache:} We use the standard uncompressed KV cache as the accuracy performance upper bound and system efficiency lower bound.
    \item \textbf{ASVD (example-based)}~\cite{asvd}: ASVD performs activation-aware SVD and compresses it at the granularity of an example for each incoming query.
    \item \textbf{Palu (global-based)}~\cite{palu}: Palu employs a low-rank projection matrix and applies KV cache compression to the global context.
\end{itemize}

\renewcommand{\arraystretch}{1.1}
\begin{table*}[!t]
\centering
\caption{
Accuracy comparison of \ourmethod against baseline methods on various many-shot ICL tasks.
All experiments are conducted using 512-shot examples under the same rank budget of 50\%.
}
\vspace{-2mm}
\label{tab:main_results}
\resizebox{\linewidth}{!}{
\begin{tabular}{l|r||cccccc || c} 
\toprule
\textbf{Model} & \textbf{Method} & BANKING77 & CLINIC150 & TREC & NLU & MATHQA & SVAMP & Avg.\\
\midrule
\multirow{4}{*}{Qwen2.5-3B} 
  & Full Cache  & 0.772 & 0.736 & 0.884 & 0.804 & 0.328 & 0.536 & 0.677 \\
  & ASVD   & 0.722 & 0.694 & 0.812 & 0.723 & 0.241 & 0.441 & 0.605 \\
  & Palu   & 0.740 & 0.712 & 0.856 & 0.768 & 0.264 & 0.484 & 0.637 \\
  & \textbf{\ourmethod} & 0.744 & 0.720 & 0.860 & 0.780 & 0.276 & 0.500 & 0.647 \\
\midrule
\multirow{4}{*}{Qwen2.5-7B} 
  & Full Cache  & 0.812 & 0.808 & 0.932 & 0.860 & 0.420 & 0.684 & 0.753 \\
  & ASVD  & 0.785 & 0.742 & 0.837 & 0.815 & 0.311 & 0.594 & 0.680 \\
  & Palu  & 0.802 & 0.763 & 0.861 & 0.824 & 0.356 & 0.592 & 0.700 \\
  & \textbf{\ourmethod} & 0.804 & 0.794 & 0.894 & 0.834 & 0.394 & 0.632 & 0.725 \\
\bottomrule
\end{tabular}
}
\end{table*}
\minisection{Implementation Details.}
We conduct our experiments on a single NVIDIA L4-24\,GB GPU, with a 2.60\, GHz Intel Xeon CPU with 512\,GB DRAM and PCIe Gen-4.
Our method is implemented on top of FlexGen~\cite{flexgen}, Transformers~\cite{transformers}, PyTorch~\cite{pytorch}, and Triton~\cite{triton}.
In terms of selecting demonstration examples, we follow prior works~\cite{dbsa,adapshot} and use the standard BM25 retriever~\cite{bm25}.

\begin{figure*}[!t]
    \centering
    \includegraphics[width=\linewidth]{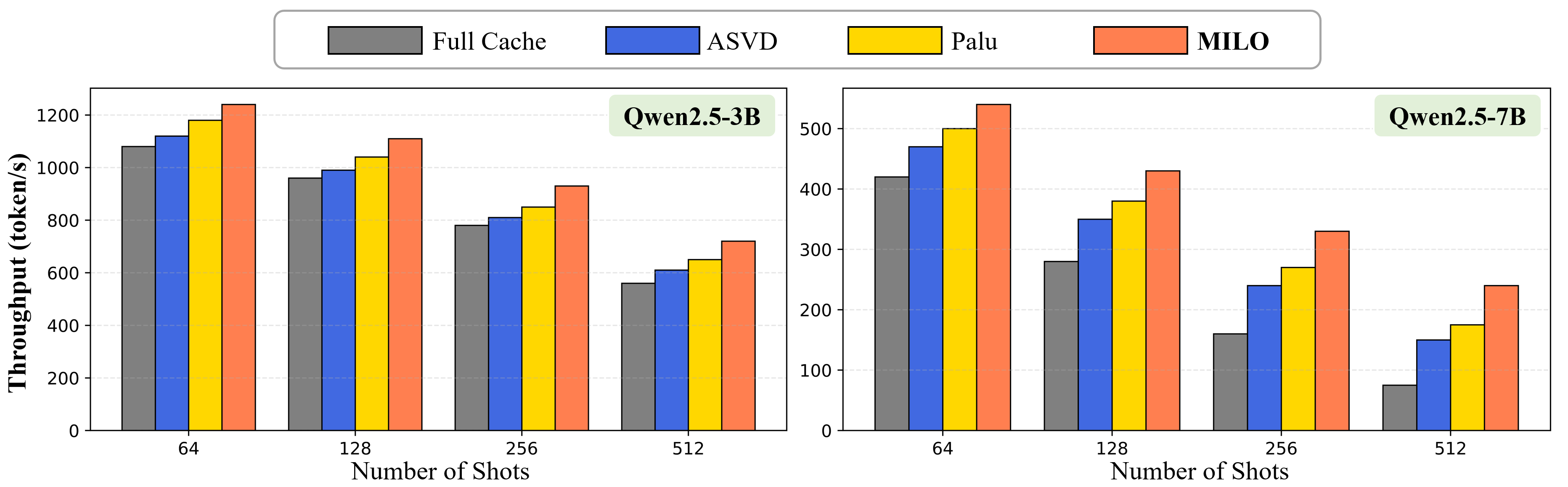}
    \vspace{-7mm}
    \caption{End-to-end system performance comparison of \ourmethod against prior methods on the MathQA dataset under the same rank budget of 50\% on a single NVIDIA L4 GPU. 
    }
    \label{fig:perf}
\end{figure*}

\subsection{Main Results}
\minisection{Accuracy Results.}
Table~\ref{tab:main_results} reports the detailed accuracy results of our method and baseline methods.
Under an identical rank budget constraint, our method outperforms prior state-of-the-art low-rank methods, improving the accuracy performance by up to 4.5\%.
Specifically, on Qwen2.5-3B, \ourmethod demonstrates a consistent lead across all individual tasks, substantially narrowing the empirical performance gap to the accuracy performance upper bound.
A similar trend can be observed on the larger Qwen2.5-7B model, where \ourmethod retains a greater proportion of the full cache representation capacity, significantly outperforming SOTA methods.

\minisection{Throughput Results.}
We further evaluate the system performance of \ourmethod under a memory-constrained deployment. 
Since the KV cache often cannot remain resident in GPU memory, we consider a KV offloading configuration where pre-encoded cache blocks for many-shot examples are stored in CPU memory and transferred to the GPU on demand during query processing~\cite{flexgen,alisa}.
Figure~\ref{fig:perf} reports the end-to-end throughput comparison.
Here, we have two key observations.
First, compared with full KV offloading, \ourmethod achieves an average speedup of 1.6$\times$, thanks to the reduced KV volume by performing low-rank compression and efficient kernel implementation.
For instance, for the 7B model with 512-shot inputs, \ourmethod can sustain up to 3.2$\times$ more throughput.
Second, \ourmethod outperforms Palu and ASVD by up to 37\% and 60\%, respectively.
Third, the larger the model size, the higher the improvement \ourmethod brings.
This is due to the increased KV cache size transferred during offload, thus making the inference more memory-bound.

\subsection{Performance Analysis}

\begin{figure*}[!t]
    \centering
    \includegraphics[width=\linewidth]{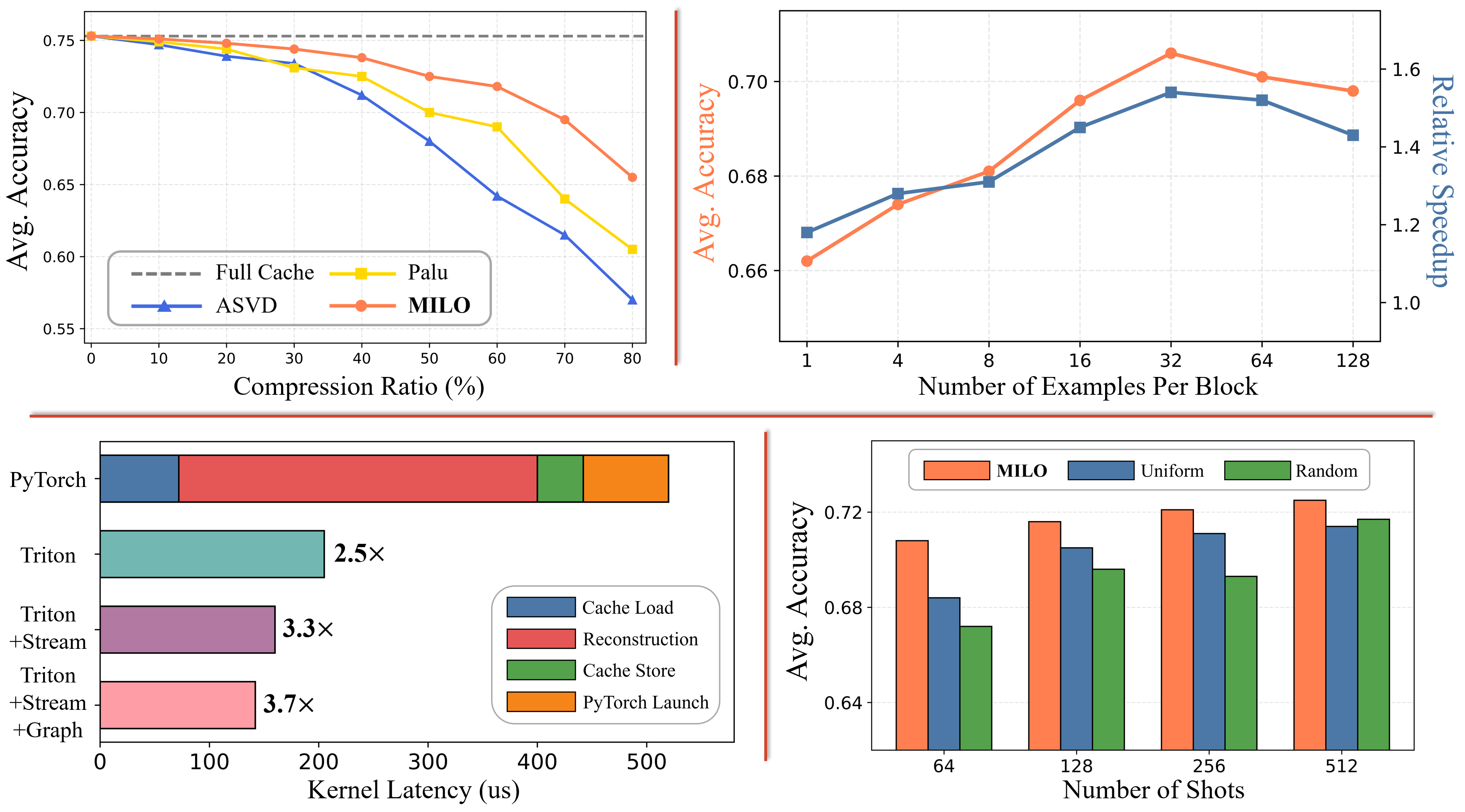}
    \vspace{-7mm}
    \caption{Performance Analysis. 
    Top Left (a): Impact of compression ratio (\%) on model performance for \ourmethod and baseline methods;
    Top Right (b): Impact of block size (number of examples per block) on latency and model accuracy for \ourmethod.;
    Bottom Left (c): Performance breakdown for \ourmethod customized Triton kernels;
    Bottom Right (d): Impact of rank allocation strategy on model accuracy for \ourmethod.
    All experiments are conducted using Qwen2.5-7B.
    }
    \label{fig:ab1}
\end{figure*}

\minisection{Impact of Rank Budget and Block Size.}
Figure~\ref{fig:ab1} (a) and (b) demonstrate the sensitivity of \ourmethod to the compression ratio and block size (number of examples per block). 
Here, we have two key observations.
First, as the compression ratio increases, all low-rank methods experience gradual accuracy degradation due to the reduced rank budget. 
However, \ourmethod consistently maintains higher accuracy than ASVD and Palu across the entire range. 
This indicates that the proposed block-wise adaptive allocation preserves more task-relevant information under constrained rank budgets.
Second, increasing the block size initially improves both model accuracy and system efficiency, as larger blocks expose more cross-example redundancy that can be captured by the low-rank representation. 
However, both accuracy and speedup begin to decline after a certain block size (32), suggesting that overly large blocks reduce compression and retrieval granularity while increasing reconstruction overhead.

\minisection{Kernel Performance Analysis.}
We further evaluate the effectiveness of the system optimizations for \ourmethod, as shown in Figure~\ref{fig:ab1} (c).
The native PyTorch implementation incurs substantial latency from separate cache loading, low-rank reconstruction, cache storage, and repeated kernel launches. 
Replacing these operations with our fused Triton implementation reduces kernel latency by approximately $2.5\times$, primarily by eliminating intermediate HBM traffic. 
Enabling parallel block reconstruction with CUDA streams further improves the speedup to approximately $3.3\times$ and combining with CUDA graph achieves a total speedup of approximately $3.7\times$ over the PyTorch implementation. 

\minisection{Impact of Rank Allocation.}
We also compare the proposed entropy-based rank allocation strategy against uniform and random allocation under the same total rank budget in Figure~\ref{fig:ab1}~(d).
We can see that \ourmethod consistently achieves the highest model accuracy. 
As the number of shots increases, the performance gap narrows, but \ourmethod continues to provide the strongest accuracy, demonstrating that adaptive rank allocation makes more effective use of a fixed compression budget.

\begin{figure*}[!t]
\noindent
\begin{minipage}[t]{0.44\textwidth}
\vspace{0pt}

\minisection{Combining KV Quantization.}
Here, we further apply asymmetric INT8~\cite{flexgen} and KIVI~\cite{kivi} quantization on top of \ourmethod and evaluate the impact on model accuracy and KV cache memory reduction ratio.
Table~\ref{tab:quantization} shows that \ourmethod can be effectively combined with quantization methods, providing complementary memory savings while preserving model accuracy.
\end{minipage}
\hfill
\begin{minipage}[t]{0.52\textwidth}
\vspace{0pt}
\centering
\captionof{table}{
Accuracy and KV-cache memory reduction results when combining
\ourmethod with quantization.
}
\label{tab:quantization}
\vspace{-3mm}
\resizebox{.8\linewidth}{!}{
\begin{tabular}{r|r||c||c}
\toprule
\textbf{Model} &
\textbf{Method} &
\textbf{KV Reduction} &
\textbf{Accuracy} \\
\midrule
\multirow{4}{*}{Qwen2.5-3B}
  & Full Cache & 1$\times$ & 0.677 \\
  & \ourmethod & 2$\times$ & 0.647 \\
  & \ourmethod + INT8 & 4$\times$ & 0.621 \\
  & \ourmethod + KIVI & 4.7$\times$ & 0.608 \\
\midrule
\multirow{4}{*}{Qwen2.5-7B}
  & Full Cache & 1$\times$ & 0.753 \\
  & \ourmethod  & 2$\times$ & 0.725 \\
  & \ourmethod + INT8 & 4$\times$ & 0.712 \\
  & \ourmethod + KIVI & 5.3$\times$ & 0.701 \\
\bottomrule
\end{tabular}
}
\end{minipage}
\end{figure*}
\vspace{-12pt}
\section{Conclusion}
\label{sec:conclusion}
In this work, we present \ourmethod, a block-wise low-rank KV-cache compression framework for efficient many-shot in-context learning. 
By exploiting heterogeneous low-rank redundancy across demonstration blocks and dynamically allocating a global rank budget, \ourmethod reduces KV cache memory footprint by 50\% while preserving model accuracy. 
Combining with customized fused Triton kernels, \ourmethod further improves the system performance, achieving up 1.8$\times$ speedup upon existing baseline methods, demonstrating a practical path toward efficient many-shot ICL inference.

\bibliography{main}

\end{document}